\documentclass[runningheads]{llncs}
\usepackage{graphicx}
\usepackage{amsmath,amssymb} 
\usepackage{color}
\usepackage[width=122mm,left=12mm,paperwidth=146mm,height=193mm,top=12mm,paperheight=217mm]{geometry}
\usepackage{times}
\usepackage{epsfig}
\usepackage{subfigure}
\usepackage{algorithm}
\usepackage{algorithmicx}
\usepackage{algpseudocode}
\usepackage[noadjust]{cite}
\usepackage{array}
\usepackage{wrapfig}
\usepackage{url}
\usepackage{standalone}

\newcolumntype{C}[1]{>{\centering\let\newline\\\arraybackslash\hspace{0pt}}m{#1}}

\newcommand{\ie}{\mbox{\emph{i. e.\ }}}
\newcommand{\eg}{\mbox{\emph{e. g.\ }}}
\let\bs = \mathbf

\let\set = \mathcal

\begin{document}
\pagestyle{headings}
\mainmatter

\title{Capturing Dynamic Textured Surfaces \\of Moving Targets} 

\titlerunning{Capturing Dynamic Textured Surfaces of Moving Targets}

\authorrunning{R. Wang et al.}

\author{Ruizhe Wang\textsuperscript{1}, 
        Lingyu Wei\textsuperscript{1}, 
        Etienne Vouga\textsuperscript{2}, \\
        Qixing Huang\textsuperscript{3},
        Duygu Ceylan\textsuperscript{4},
        G\'{e}rard Medioni\textsuperscript{1},
        and Hao Li\textsuperscript{1}}
\institute{\textsuperscript{1}University of Southern California \\
           \url{{ruizhewa, lingyu.wei, medioni}@usc.edu} \\
           \url{hao@hao-li.com} \\
           \textsuperscript{2}University of Texas at Austin \\
           \url{evouga@cs.utexas.edu} \\
           \textsuperscript{3}Toyota Technological Institute at Chicago \\
           \url{huangqx@ttic.edu} \\
           \textsuperscript{4}Adobe Research \\
           \url{ceylan@adobe.com}}

\maketitle

\begin{abstract}

We present an end-to-end system for reconstructing complete watertight and textured models of moving subjects such as clothed humans and animals, using only three or four handheld sensors. The heart of our framework is a new pairwise registration algorithm that minimizes, using a particle swarm strategy, an alignment error metric based on mutual visibility and occlusion. We show that this algorithm reliably registers partial scans with as little as 15\% overlap without requiring any initial correspondences, and outperforms alternative global registration algorithms. This registration algorithm allows us to reconstruct moving subjects from free-viewpoint video produced by consumer-grade sensors, without extensive sensor calibration, constrained capture volume, expensive arrays of cameras, or templates of the subject geometry.


\end{abstract}

\section{Introduction}
\label{sec:intro}

The rekindling of interest in immersive, 360-degree virtual environments, spurred on by the Oculus, Hololens, and other breakthroughs in consumer AR and VR hardware, has birthed a need for digitizing objects with full geometry and texture from all views. One of the most important objects to digitize in this way are moving, clothed humans, yet they are also among the most challenging: the human body can undergo large deformations over short time spans, has complex geometry with occluded regions that can only be seen from a small number of angles, and has regions like the face with important high-frequency features that must be faithfully preserved.

Most techniques for capturing high-quality digital humans rely on a large array of sensors mounted around a fixed capture volume. The recent work of Collet et al.~\cite{Collet:2015} uses such a setup to capture live performances and compresses them to enable streaming of free-viewpoint videos. Unfortunately, these techniques are severely restrictive: first, to ensure high-quality reconstruction and sufficient coverage, a large number of expensive sensors must be used, leaving human capture out of reach of consumers without the resources of a professional studio. Second, the subject must remain within the small working volume enclosed by the sensors, ruling out subjects interacting with large, open environments or undergoing large motions.

Using free-viewpoint sensors is an attractive alternative, since it does not constrain the capture volume and allows ordinary consumers, with access to only portable, low-cost devices, to capture human motion. The typical challenge with using hand-held active sensors is that, obviously, multiple sensors must be used simultaneously from different angles to achieve adequate coverage of the subject. In overlapping regions, signal interference causes significant deterioration in the quality of the captured geometry. This problem can be avoided by minimizing the amount of overlap between sensors, but on the other hand, existing registration algorithms for aligning the captured partial scans only work reliably if the partial scans significantly overlap. Template-based methods like the work of Ye et al~\cite{Ye:2013} circumvent these difficulties by warping a full geometric template to track the moving sparse partial scans, but templates are only readily available for naked humans~\cite{SCAPE}; for clothed humans a template must be precomputed on a case-by-case basis.

We thus introduce a new shape registration method that can reliably register partial scans even with \emph{almost no overlap}, sidestepping the need for shape templates or sensor arrays. This method is based on a \emph{visibility error metric} which encodes the intuition that if a set of partial scans are properly registered, each partial scan, when viewed from the same angle at which it was captured, should occlude all other partial scans. We solve the global registration problem by minimizing this error metric using a particle swarm strategy, to ensure sufficient coverage of the solution space to avoid local minima. This registration method significantly outperforms state of the art global registration techniques like 4PCS~\cite{aiger20084} for challenging cases of small overlap.

\paragraph{Contributions.} We present the first end-to-end free-viewpoint reconstruction framework that produces watertight, fully-textured surfaces of moving, clothed humans using only three to four handheld depth sensors, without the need of shape templates or extensive calibration. The most significant technical component of this system is a robust pairwise global registration algorithm, based on minimizing a visibility error metric, that can align depth maps even in the presence of very little (15\%) overlap.

\section{Related Work}
\label{sec:related}

Digitizing realistic, moving characters has traditionally involved an intricate pipeline including modeling, rigging, and animation. 
This process has been occasionally assisted by 3D motion and geometry capture systems such as marker-based motion capture or markerless capture methods involving large arrays of sensors~\cite{debevec_light_2012}. Both approaches supply artists with accurate reference geometry and motion, but they require specialized hardware and a controlled studio setting.

Real-time 3D scanning and reconstruction systems requiring only a single sensor, like KinectFusion~\cite{Izadi:2011}, allow casual users to easily scan everyday objects; however, as with most simultaneous localization and mapping (SLAM) techniques, the major assumption is that the scanned scene is rigid. This assumption is invalid for humans, even for humans attempting to maintain a single pose; several follow-up works have addressed this limitation by allowing near-rigid motion, and using non-rigid partial scan alignment algorithms~\cite{Tong:2012,Li:2013}. While the recent DynamicFusion framework~\cite{Newcombe_2015_CVPR} and similar systems~\cite{Dou:2015} show impressive results in capturing non-rigidly deforming scenes, our goal of capturing and tracking freely moving targets is fundamentally different: we seek to reconstruct a \emph{complete} model of the moving target at all times, which requires either extensive prior knowledge of the subject's geometry, or the use of multiple sensors to provide better coverage.

Prior work has proposed various simplifying assumptions to make the problem of capturing entire shapes in motion tractable. Examples include assuming availability of a template, high-quality data, smooth motion, and a controlled capture environment.

\paragraph{Template-based Tracking:} The vast majority of related work on capturing dynamic motion focuses on specific human parts, such as faces~\cite{Li:2013:RFA} and hands~\cite{qian2014realtime,oikonomidis2012tracking}, for which specialized shapes and motion templates are available. In the case of tracking the full human body, parameterized body models~\cite{Bogo:ICCV:2015} have been used. However, such models work best on naked subjects or subjects wearing very tight clothing, and are difficult to adapt to moving people wearing more typical garments.

Another category of methods first capture a template in a static pose and then track it across time. Vlasic et al~\cite{Vlasic:2008} use a rigged template model, and De Aguiar et al~\cite{deAguiar:2008} apply a skeleton-less shape deformation model to the template to track human performances from multi-view video data. Other methods~\cite{Li:2009,Zollhofer:2014} use a smoothed template to track motion from a capture sequence. The more recent work of Wu et al.~\cite{Wu:2013} and Liu et al.~\cite{Liu2014} track both the surface and the skeleton of a template from stereo cameras and sparse set of depth sensors respectively.

All of these template-based approaches handle with ease the problem of tracking moving targets, since the entire geometry of the target is known. However, in addition to requiring constructing or fitting said template, these methods share the common limitation that they cannot handle geometry or topology changes which are likely to happen during typical human motion (picking up an object; crossing arms; etc).

\paragraph{Dynamic Shape Capture:} Several works have proposed to reconstruct both shape and motion from a dynamic motion sequence. 
Given a series of time-varying point clouds, Wand et al.~\cite{Wand:2007} use a uniform deformation model to capture both geometry and motion. A follow-up work~\cite{Wand:2009:ERN} proposes to separate the deformation models used for geometry and motion capture. Both methods make the strong assumption that the motion is smooth, and thus suffer from popping artifacts in the case of large motions between time steps. S\"{u}{\ss}muth et al.~\cite{Sussmuth:2008} fit a 4D space-time surface to the given sequence but they assume that the complete shape is visible in the first frame. Finally, Tevs et al.~\cite{Tevs:2012} detect landmark correspondences which are then extended to dense correspondences. While this method can handle a considerable amount of topological change, it is sensitive to large acquisition holes, which are typical for commercial depth sensors. 

Another category of related work aims to reconstruct a deforming watertight mesh from a dynamic capture sequence by imposing either visual hull~\cite{Vlasic:2009} or temporal coherency constraints~\cite{Li:2012:TCC:2077341.2077343}. Such constraints either limit the capture volume or are not sufficient to handle large holes. Furthermore, neither of these methods focus on propagating texture to invisible areas; in contrast, 
we use dense correspondences to perform texture inpainting in non-visible regions. Bojsen-Hansen et al.~\cite{TSwET_2012} also use dense correspondences to track surfaces with evolving topologies. However, their method requires the input to be a closed manifold surface. Our goal, on the other hand, is to reconstruct such complete meshes from sparse partial scans.

%
The recent work of Collet et al.~\cite{Collet:2015} uses multimodal input data from a stage setup to capture topologically-varying scenes. While this method produces impressive results, it requires a pre-calibrated complex setup. In contrast, we use a significantly cheaper and more convenient setup composed of three to four commercial depth sensors.

\paragraph{Global Range Image Registration:} At the heart of our approach is a robust algorithm that registers noisy data coming from each commercial depth sensor with very little overlap. A typical approach is to first perform global registration to compute an approximate rigid transformation between a pair of range images, which is then used to initialize local registration methods (e.g., Iterative Closest Point (ICP) \cite{zhang1994iterative,chen1991object}) for further refinement. A popular approach for global registration is to construct feature descriptors for a set of interest points which are then correlated to estimate a rigid transformation. Spin-images~\cite{johnson1999using}, integral volume descriptors~\cite{gelfand2005robust}, and point feature histograms (PFH, FPFH)~\cite{rusu2008aligning,rusu2009fast} are among the popular descriptors proposed by prior work.  Makadia et al.~\cite{makadia2006fully} represent each range image as a translation-invariant emph{extended gaussian Image (EGI)}~\cite{horn1984extended} using surface normals. They first compute the optimum rotation by correlating two EGIs and further estimate the corresponding translation using Fourier transform. For noisy data as coming from a commercial depth sensor, however, it is challenging to compute reliable feature descriptors. Another approach for global registration is to align either main axes extracted by principal component analysis (PCA)~\cite{chung1998registration} or a sparse set of control points in a RANSAC loop~\cite{chen1999ransac}. Silva et al.~\cite{silva2005precision} introduce a robust \emph{surface interpenetration measure (SIM)} and search the 6 DoF parameter space with a genetic algorithm. More recently, Yang et al.~\cite{yang2013go} adopt a branch-and-bound strategy to extend the basic ICP algorithm in a global manner. 4PCS \cite{aiger20084} and its latest variant Super-4PCS \cite{mellado2014super} register a pair of range images by extracting all coplanar 4-points sets. Such approaches, however, are likely to converge to wrong alignments in cases of very little overlap between the range images (see Section~\ref{sec:regResults}). 
%

Several prior works have adopted silhouette-based constraints for aligning multiple images~\cite{580392,Matusik:2000:IVH,4155753,4587696,Starck:2007:SCP,Wu_ICCV2011,hernandez2007silhouette}. While the idea is similar to our approach, our registration algorithm also takes advantage of depth information, and employs a particle-swarm optimization strategy that efficiently explores the space of alignments.

\section{System Overview}
\label{sec:overview}

\begin{figure}[t]
 \centering
\includegraphics[width=\textwidth]{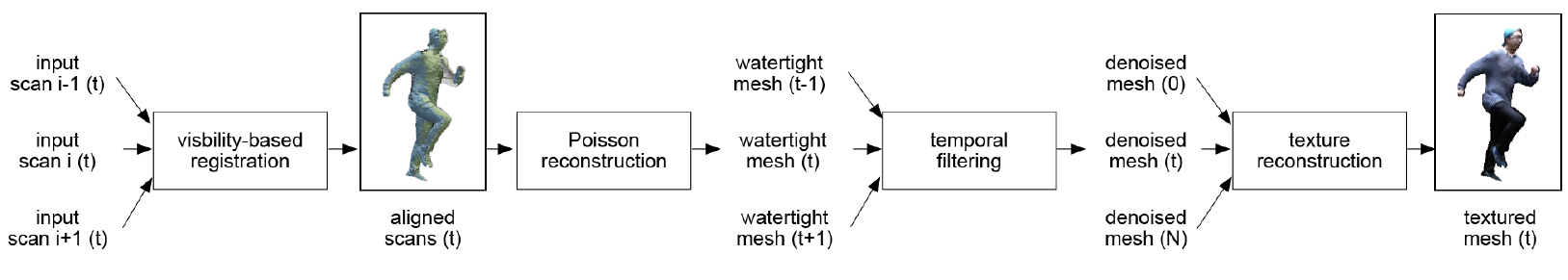}
\caption{An overview of our textured dynamic surface capturing system.
 \label{fig:overview}}
\end{figure}

Our pipeline for reconstructing fully-textured, watertight meshes from three to four depth sensors can be decomposed into four major steps. See Figure~\ref{fig:overview} for an overview of how these steps interrelate.


\emph{1. Data Capture:} We capture the subject (who is free to move arbitrarily) using uncalibrated hand-held real-time RGBD sensors. We experimented with both Kinect One time-of-flight cameras mounted on laptops, and Occipital Structure IO sensors mounted on iPad Air 2 tablets (section~\ref{sec:results}).

\emph{2. Global Rigid Registration:} The relative positions of the depth sensors constantly change over time, and the captured depth maps often have little overlap (10\%-30\%). For each frame, we globally register sparse depth images from all views (section~\ref{sec:registration}). This step produces registered, but incomplete, textured partial scans of the subject for each frame.


\emph{3. Surface Reconstruction:} To reduce flickering artifacts, we adopt the shape completion pipeline of Li et al~\cite{Li:2012:TCC:2077341.2077343} to warp partial scans from temporally-proximate frames to the current frame geometry. A weighted Poisson reconstruction step then extracts a single watertight surface. There is no guarantee, however, that the resulted fused surface has complete texture coverage (and indeed typically texture will be missing at partial scan seams and in occluded regions.)



\emph{4. Dense Correspondences for Texture Reconstruction:} 
We complete regions of missing or unreliable texture on one frame by propagating data from other (perhaps very temporally-distant) frames with reliable texture in that region. We adopt a recently-proposed correspondence computation framework~\cite{WeiHCVL15} based on a deep neural network to build dense correspondences between any two frames, even if the subject has undergone large relative deformations. Upon building dense correspondences, we transfer texture from reliable regions to less reliable ones. 

We next describe the details of the global registration method as it constitutes the core of our pipeline. Please refer to the supplementary material for more details of the other components.


\section{Robust Rigid Registration}
\label{sec:registration}

The key technical challenge in our pipeline is registering a set of depth images accurately without assuming any initialization, even when the geometry visible in each depth image has very little overlap with any other depth image. We attack this problem by developing a robust pairwise global registration method: let $P_1$ and $P_2$ be partial meshes generated from two depth images captured simultaneously. We seek a global Euclidean transformation $T_{12}$ which aligns $P_2$ to $P_1$. Traditional pairwise registration based on finding corresponding points on $P_1$ and $P_2$, and minimizing the distance between them, has notorious difficulty in this setting. As such we propose a novel \emph{visibility error metric} (VEM) (Section~\ref{sec:VEM}), and we minimize the VEM to find $T_{12}$  (Section~\ref{sec:xform}). We further extend this pairwise method to handle multi-view global registration (Section~\ref{sec:Multi-view}).


\subsection{Visibility Error Metric}
\label{sec:VEM}

\begin{wrapfigure}{r}{0.50\textwidth}
\centering
\includegraphics[width=0.48\textwidth]{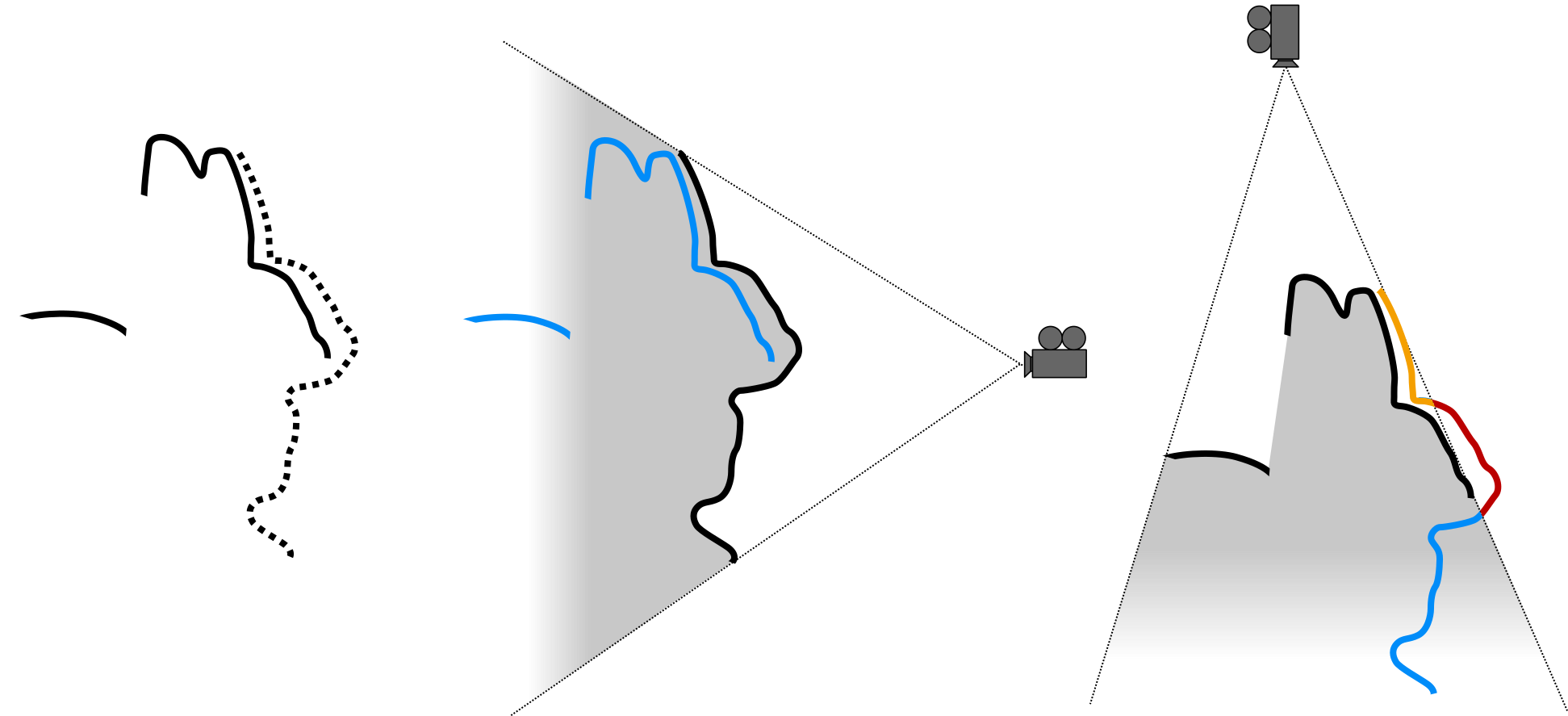}
\caption{Left: two partial scans $P_1$ (dotted) and $P_2$ (solid) of a 2D bunny. Middle: when viewed from $P_1$'s camera, $P_2$ is entirely occluded (blue). Therefore all of $P_2$ is in $\mathcal{O}$. Right: when viewed from $P_2$'s camera, parts of $P_1$ are in $\mathcal{O}$ (blue), parts occlude $P_2$ and are thus in $\mathcal{F}$ (yellow), and parts are in $\mathcal{B}$ (red).}
\label{fig:metric}
\end{wrapfigure}

Suppose $P_1$ and $P_2$ are correctly aligned, and consider looking at the pair of scans through a camera whose position and orientation matches that of the sensor used to capture $P_1$. The only parts of $P_2$ that should be visible from this view are those that overlap with $P_1$: parts of $P_2$ that do not overlap should be completely occluded by $P_1$ (otherwise they would have been detected and included in $P_1$). Similarly, when looking at the scene through the camera that captured $P_2$, only parts of $P_1$ that overlap with $P_2$ should be visible.

\emph{Visibility-Based Alignment Error} We now formalize the above idea. Let $P_1, P_2$ be two partial scans, with $P_1$ captured using a sensor at position $c_p$ and view direction $c_v$. For every point $x\in P_2$, let $I(x)$ be the first intersection point of $P_1$ and the ray $\overrightarrow{c_px}$. We can partition $P_2$ into three regions, and associate to each region an energy density $d(x,P_1)$ measuring the extent to which points $x$ in that region violate the above visibility criteria:
\begin{itemize}
\item points $x\in \mathcal{O}$ that are occluded by $P_1$: $\|x-c_p\| \geq \|I(x)-c_p\|.$ To points in this region we associate no energy:
$$d_{\mathcal{O}}(x,P_1) = 0.$$
\item points $x\in \mathcal{F}$ that are in front of $P_1$: $\|x-c_p\| < \|I(x)-c_p\|.$ Such points might exist even when $P_1$ and $P_2$ are well-aligned, due to surface noise and roughness, etc. However, we penalize large violations using:
$$d_{\mathcal{F}}(x,P_1) = \|x-I(x)\|^2.$$
\item points $x\in \mathcal{B}$ for which $I(x)$ does not exist. Such points also violate the visibility criteria. It is tempting to penalize such points proportionally to the distance between $x$ and its closest point on $P_1$, but a small misalignment could create a point in $\mathcal{B}$ that is very distant from $P_1$ in Euclidean space, despite being very close to $P_1$ on the camera image plane. We therefore penalize $x$ using squared distance on the image plane,
$$d_{\mathcal{B}}(x,P_1) = \min_{y\in S_1} \left\|\set{P}_{c_v}x - \set{P}_{c_v} y\right\|^2,$$
where $\set{P}_{c_v}$ is the projection $I-c_vc_v^T$ onto the plane orthogonal to $c_v$.
\end{itemize}

Figure~\ref{fig:metric} illustrates these regions on a didactic 2D example. Alignment of $P_1$ and $P_2$ from the point of view of $P_1$ is then measured by the aggregate energy $d(P_2,P_1) = \sum_{x\in P_2} d(x,P_1)$. Finally, every Euclidean transformation $T_{12}$ that produces a possible alignment between $P_1$ and $P_2$ can be associated with an energy to define our visibility error metric on $SE(3)$,
\begin{equation}
E(T_{12}) = d\left(T_{12}^{-1}P_1, P_2\right) + d\left(T_{12}P_2, P_1\right) \label{eq:metric}.
\end{equation}

\subsection{Finding the Transformation}\label{sec:xform}

\begin{figure}[h]
\centering
\subfigure[]
{
\includegraphics[width=0.45\textwidth]{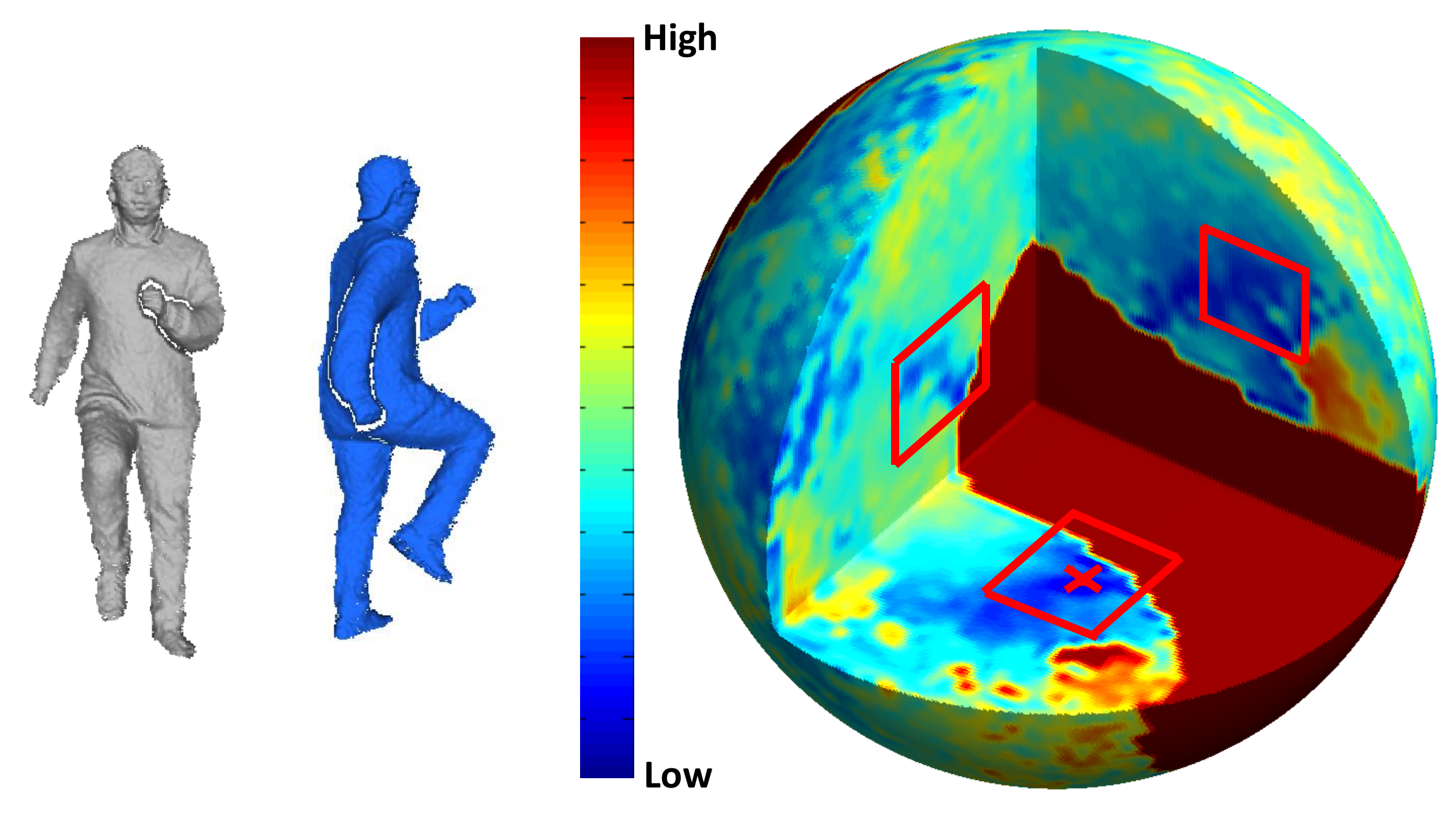}
\label{fig:FIG4}
}
\subfigure[]
{
\includegraphics[width=0.48\textwidth]{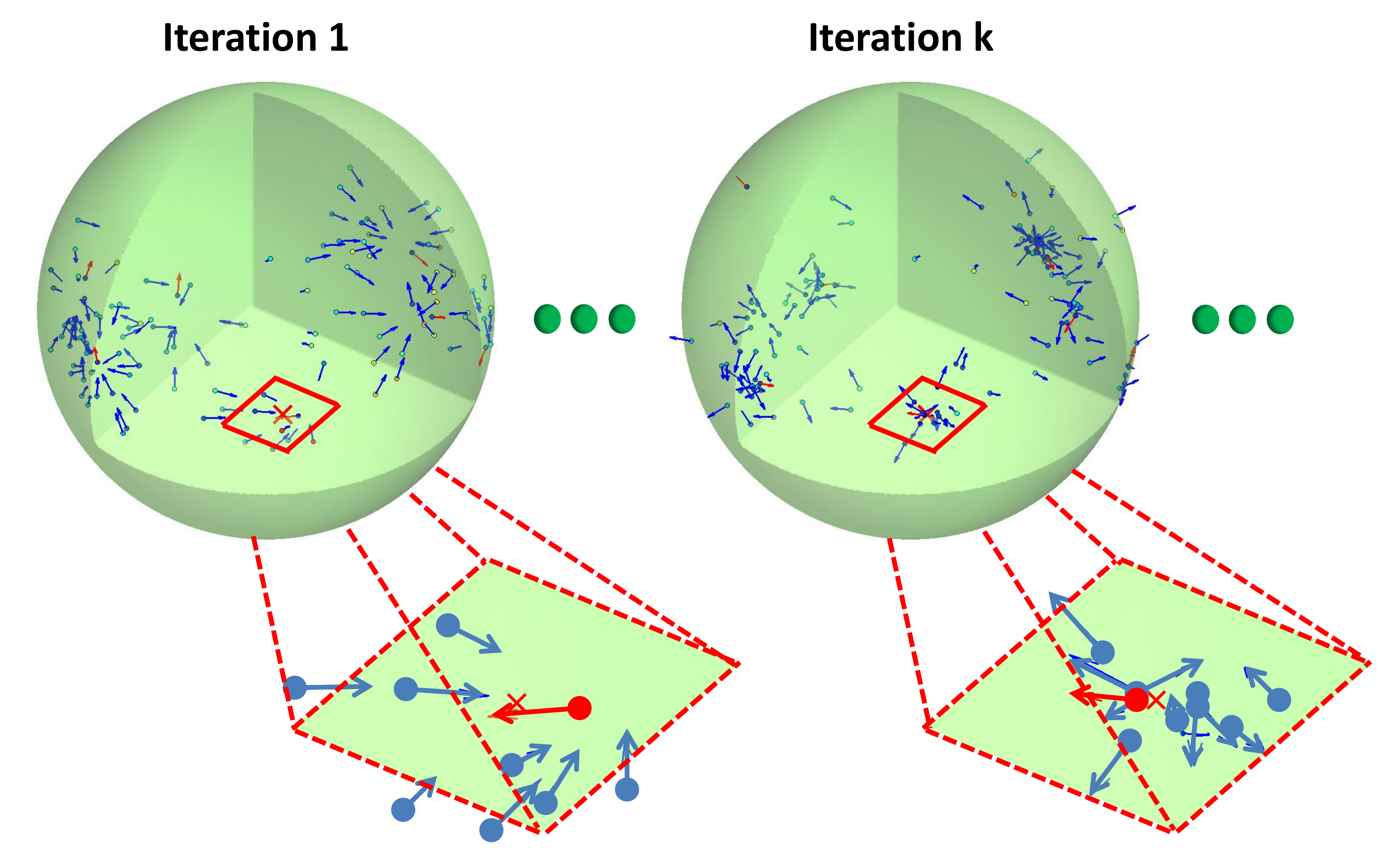}
\label{fig:FIG5}
}
\caption{(a) Left: a pair of range images to be registered. Right: VEM evaluated on the entire rotation space. Each point within the unit ball represents the vector part of a unit quaternion; for each quaternion, we estimate its corresponding translation component and evaluate the VEM on the composite transformation. The red rectangles indicate areas with local minima, and the red cross is the global minimum. (b) Example particle locations and displacements at iteration $1$ and $k$. Blue vectors indicate displacement of regular (non-guide) particles following a traditional particle swarm scheme. Red vectors are displacements of guide particles. Guide particles draw neighboring regular particles more efficiently towards local minima to search for the global minimum.}
\label{fig:FIG4-5}
\end{figure}

Minimizing the error metric~\eqref{eq:metric} consists of solving a nonlinear least squares problem and so in principle can be optimized using e.g. the Gauss-Newton method. However, it is non-convex, and prone to local minima (Figure~\ref{fig:FIG4}). Absent a straightforward heuristic for picking a good initial guess, we instead adopt a Particle Swarm Optimization (PSO)~\cite{kennedy2010particle} method to efficiently minimize~\eqref{eq:metric}, where ``particles'' are candidate rigid transformations that move towards smaller energy landscapes in $SE(3)$. We could independently minimize $E$ starting from each particle as an initial guess, but this strategy is not computationally tractable. So we iteratively update all particle positions in lockstep: a small set of the most promising \emph{guide} particles, that are most likely to be close to the global minimum, are updated using an iteration of Levenberg-Marquardt. The rest of the particles receive PSO-style weighted random perturbations. This procedure is summarized in Algorithm~\ref{SPSOAlg}, and each step is described in more detail below.

\begin{algorithm}
\caption{Modified Particle Swarm Optimization}
\label{SPSOAlg}
\begin{algorithmic}[1]
\State{\emph{Input: A set of initial ``particles'' (orientations) $\{\mathbf{T}^0_1,...,\mathbf{T}^0_N\} \in SE(3)^N$}}
\State{evaluate VEM on initial particles}
\For{each iteration}
\State{select guide particles}
\For{each guide particle}
\State{update guide particle using Levenberg-Marquardt}
\EndFor
\For{each regular particle}
\State{update particle using weighted random displacement}
\EndFor
\State{recalculate VEM at new locations}
\EndFor
\State{\emph{Output: The best particle $\mathbf{T}^b$}}
\end{algorithmic}
\end{algorithm}

\begin{wrapfigure}{r}{0.40\textwidth}
\centering
\includegraphics[width=0.38\textwidth]{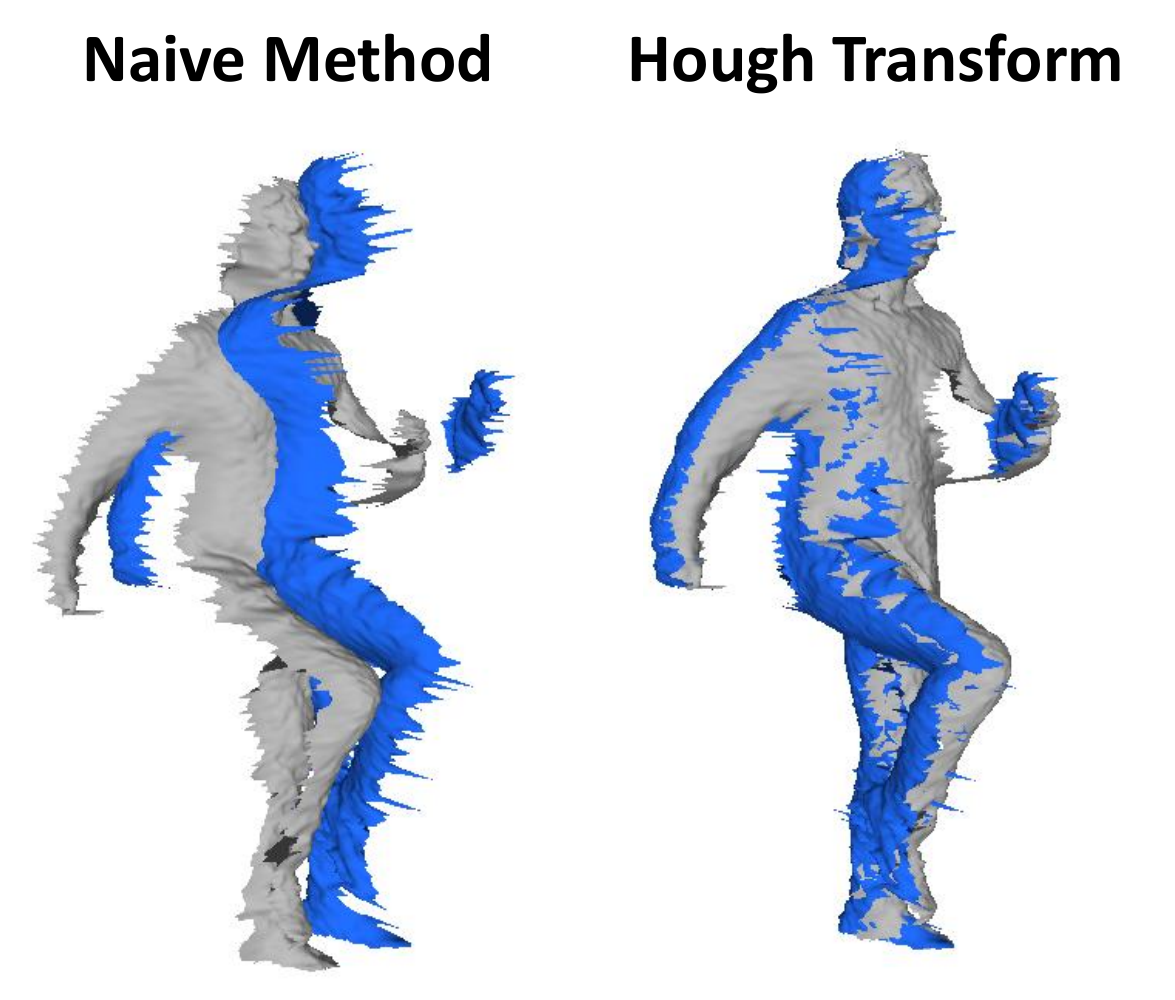}
\caption{Translation estimation examples of our Hough Transform method on range scans with limited overlap. The na\"{i}ve method, which simply aligns the corresponding centroids, fails to estimate the correct translation.}
\label{fig:FIG3}
\end{wrapfigure}

\emph{Initial Particle Sampling} We begin by sampling $N$ particles (we use $N=1600$), where each particle represents a rigid motion $m_i \in SE(3)$. Since $SE(3)$ is not compact, it is not straightforward to directly sample the initial particles. We instead uniformly sample only the rotational component $R_i$ of each particle \cite{shoemake1992uniform}, and solve for the best translation using the following Hough-transform-like procedure. For every $x\in P_1$ and $y\in R_i P_2$, we measure the angle between their respective normals, and if it is less than $20^{\circ}$, the pair $(x,y)$ votes for a translation of $y-x$. These translations are binned (we use $10\mathrm{mm}\times 10\mathrm{mm}\times 10\mathrm{mm}$ bins) and the best translation $\bs{t}_i^0$ is extracted from the bin with the most votes. The translation estimation procedure is robust even in the presence of limited overlap amount (Figure~\ref{fig:FIG3}).

The above procedure yields a set $\set{T}^0 = \{T_i^0\} = \{(R_i^0, \bs{t}_i^0)\}$ of $N$ initial particles. We next describe how to step the orientation particles from their values $\set{T}^k$ at iteration $k$ to $\set{T}^{k+1}$ at iteration $k+1$.

\emph{Identifying Guide Particles} We want to select as guide particles those particles with lowest visibility error metric; however we don't want many clustered redundant guide particles. Therefore we first promote the particle $T^k_i$ with lowest error metric to guide particle, then remove from consideration all nearby particles, e.g. those that satisfy
$$
d_{\theta}(R^k_{j}, R^k_{i}) \leq \theta_{r},
$$
where $d_{\theta}(R^k_i,R^k_j) = \theta\left(\log \left[R^k_j\right]^{-1}R^k_i\right)$ is the bi-invariant metric on $SO(3)$, e.g. the least angle of all rotations $R$ with $R^k_i = RR^k_j.$ We use $\theta_{\tau} = 30^{\circ}$. We then repeat this process (promoting the remaining particle with lowest VEM, removing nearby particles, etc) until no candidates remain.

\emph{Guide Particle Update} We update each guide particle $T^k_i$ to decrease its VEM. We parameterize the tangent space of $SE(3)$ at $T^k_i$ by two vectors $\bs{u},\bs{v} \in \mathbb{R}^3$ with $\exp (\bs{u},\bs{v}) = \left(\exp([u]_{\times})R^k_i, \bs{t}^k_i + \bs{v}\right)$, where $[u]_\times$ is the cross-product matrix. We then use the Levenberg-Marquardt method to find an energy-decreasing direction $(\bs{u},\bs{v})$, and set $T^{k+1}_i = \exp(\bs{u},\bs{v})$. Please see the supplementary material for more details.

\emph{Other Particle Update} Performing a Levenberg-Marquardt iteration on all particles is too expensive, so we move the remaining non-guide particles by applying a randomly weighted summation of each particle's displacement during the previous iteration, the displacement towards its best past position, and the displacement towards the local best particle within radius $\theta_{r}$ (measured using $d_\theta$) with lowest energy, as in standard PSO~\cite{kennedy2010particle}. While the guide particles rapidly descend to local minima, they are also local best particles and drag neighboring regular particles with them for a more efficient search of all local minima, from which the global one is extracted (Figure~\ref{fig:FIG5}). Please refer to the supplementary material for more details.

\emph{Termination} Since the VEM of each guide particle is guaranteed to decrease during every iteration, the particle with lowest energy is always selected as a guide particle, and the local minima of $E$ must lie in a bounded subset of $SE(3)$. In the above procedure the particle with lowest energy is guaranteed to converge to a local minimum of $E$. We terminate the optimization when $\min_i |E(T^{k}_i) - E(T^{k+1}_i)| \leq 10^{-4}.$ In practice this occurs within 5--10 iterations.


\subsection{Multi-view Extension}
\label{sec:Multi-view}

We extend our VEM-based pairwise registration method to globally align a total of $M$ partial scans \big\{$P_1,...,P_M$\big\} by estimating the optimum transformation set \big\{$T_{12},...,T_{1M}$\big\}. First we perform pairwise registration between all pairs to build a registration graph, where each vertex represents a partial scan and each pair of vertices are linked by an edge of the estimated transformation. We then extract all spanning trees from the graph, and for each spanning tree we calculate its corresponding transformation set \big\{$T_{12},...,T_{1M}$\big\} and estimate the overall VEM as,
\begin{equation}
E_M = \sum\limits_{i\neq j}d\left(T_{1j}^{-1}T_{1i}P_i, P_j\right) + d\left(T_{1i}^{-1}T_{1j}P_j, P_i\right)
\label{eq:overallmetric}.
\end{equation}
We select the transformation set with the minimum overall VEM. We perform several iterations of Levenberg-Marquardt algorithm to minimize Equation~\ref{eq:overallmetric} to further jointly refine the transformation set.

\emph{Temporal Coherence} When globally registering depth images from multiple sensors frame by frame, we can easily incorporate temporal coherence into the global registration framework by adding the final estimated transformation set of the previous frame to the pool of transformation sets of the current frame before selecting the best one. It is worth mentioning, however, that our capturing system does not rely on the assumption of temporal coherence and the transformation set is estimated globally for each frame. This is especially crucial for a system with handheld sensors, where the temporal coherence assumption is easily violated. 

\section{Global Registration Evaluation}
\label{sec:regResults}

\paragraph{Data Sets.} We evaluate our registration algorithm on the Stanford 3D Scanning Repository and the Princeton Shape Benchmark \cite{shilane2004princeton}. We use 4 models from the Stanford 3D Scanning Repository (the Bunny, the Happy Buddha, the Dragon, and the Amardillo), and use all 1814 models from the Princeton Shape Benchmark. We believe these two data sets, especially the latter, are general enough to cover shape variation of real world objects. For each data set, we generated 1000 pairs of synthetic depth images with uniformly varying degrees of overlap; these range maps were synthesized using randomly-selected 3D models and randomly-selected camera angles. Each pair is then initialized with a random initial relative transformation. As such, for each pair of range images, we have the ground truth transformation as well as their overlap ratio.

\paragraph{Evaluation Metric.} The extracted transformation, if not correctly estimated, can be at any distance from the ground truth transformation, depending on the specific shape of the underlying surfaces and the local minima distribution of the solution space. Thus, it is not very informative to directly use the RMSE of rotation and translation estimation. It is rather straightforward to use success percentage as the evaluation metric. We claim the global registration to be successful if the error $d_{\theta}(R_{est},R_{gt})$ of the estimated rotation $R_{est}$ is smaller than a small angle $10^{\circ}$. We do not enforce the translation to be close since it is scale-dependent and the translation component is easily recovered by a robust local registration method if the rotation component is close enough (e.g., by using surface normals to prune incorrect correspondences \cite{rusinkiewicz2001efficient}). 

\paragraph{Effectiveness of the PSO Strategy.} To demonstrate the advantage of the particle-swarm optimization strategy, we compare our full algorithm to three alternatives on the Stanford 3D Scanning Repository: 1) a baseline method that simply reports the minimum particles from all initially-sampled particles, with no attempt at optimization; 2) using only a traditional PSO formulation, without guide particles; and 3) updating only the guide particles, and applying no displacement to ordinary particles.

\begin{figure}[h]
\centering
\includegraphics[width=\columnwidth]{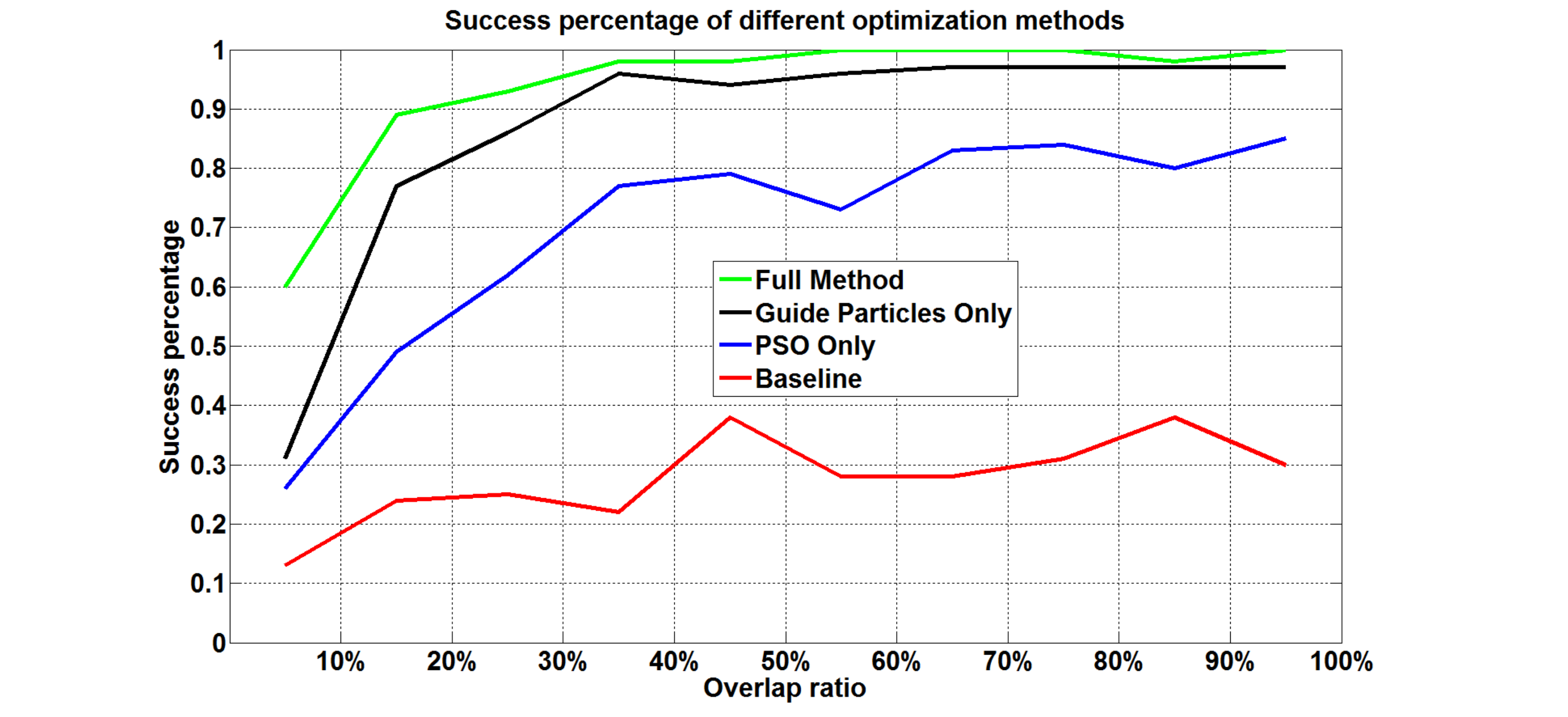}
\caption{Success percentage of the global registration method employing different optimization schemes on the Stanford 3D Scanning Repository.}
\label{fig:FIGOptimization}
\end{figure}

Figure~\ref{fig:FIGOptimization} compares the performance of the four alternatives. While updating guide particles alone achieves good registration results, incorporating the swarm intelligence further improves the performance, especially on range scans with overlap ratios below $30\%$. 

\paragraph{Comparisons.} To demonstrate the effectiveness of the proposed registration method, we compare it against four other alternatives: 1) a baseline method that aligns principal axes extracted with weighted PCA~\cite{chung1998registration}, where the weight of each vertex is proportional to its local surface area; 2) Go-ICP~\cite{yang2013go}, which combines local ICP with a branch-and-bound search to find the global minima; 3) FPFH~\cite{rusu2009fast, rusu20113d}, which matches FPFH descriptors; 4) 4PCS, a state-of-the-art method that performs global registration by constructing a congruent set of 4 points between range images \cite{aiger20084}. We do not compare with its latest variant SUPER-4PCS \cite{mellado2014super} as only efficiency is improved for the latter. 
For Go-ICP, FPFH and 4PCS, we use the authors' original implementation and tune parameters to achieve optimum performance.

\begin{figure}[h]
\centering
\includegraphics[width=\columnwidth]{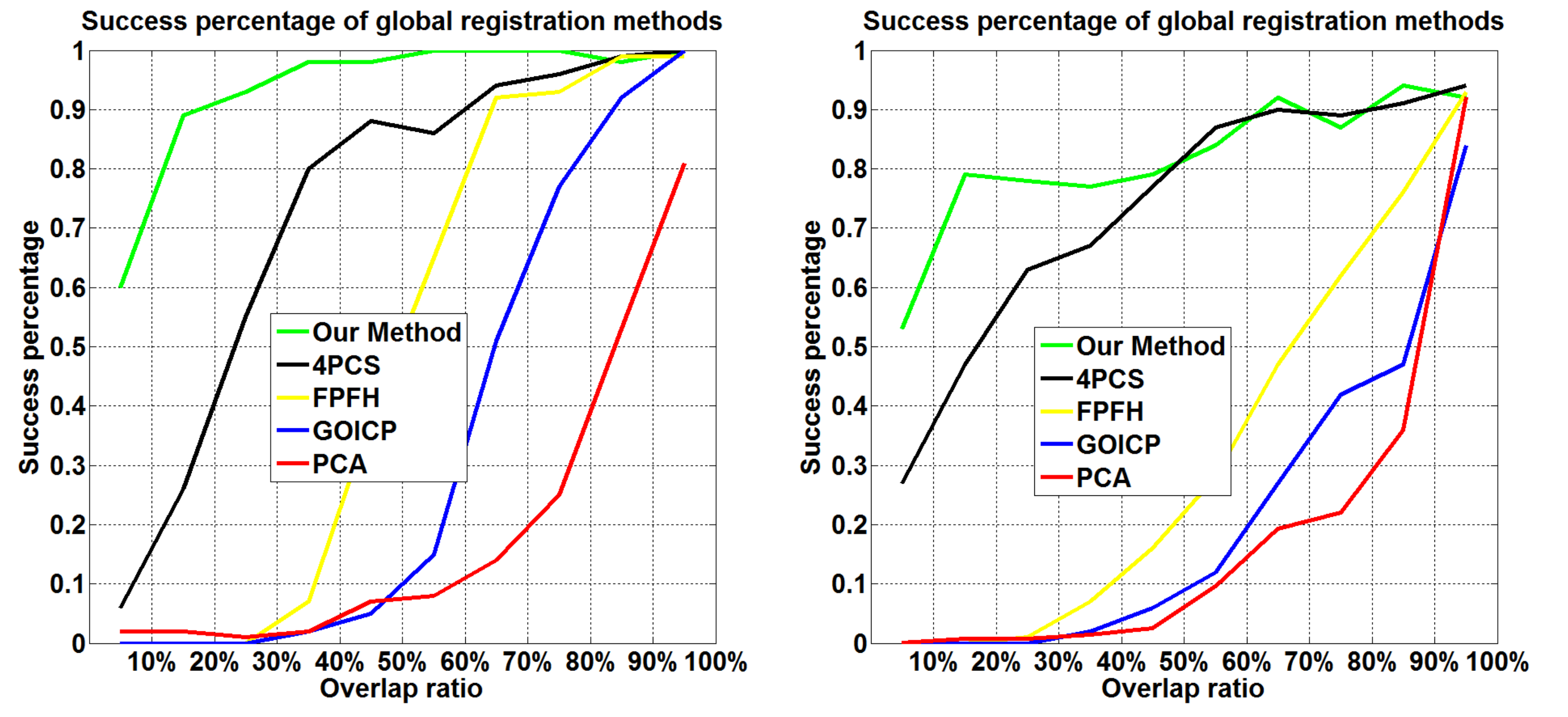}
\caption{Success percentage of our global registration method compared with other methods. Left: Comparison on the Stanford 3D Scanning Repository. Right: Comparison on the Princeton Shape Benchmark.}
\label{fig:FIGRegistration}
\end{figure}

Figure~\ref{fig:FIGRegistration} compares the performance of the five methods on the two data sets respectively. The overall performance on the Princeton Shape Benchmark is lower as this data set is more challenging with many symmetric objects. As expected the baseline PCA method only works well when there is sufficient overlap. All previous methods experience a dramatic fall in accuracy once the overlap amount drops below $40\%$; 4PCS performs the best out of these, but because 4PCS is essentially searching for the most consistent area shared by two shapes, for small overlap ratio, it can converge to false alignments (Figure~\ref{fig:FIGRegistrationExamples}). Our method outperforms all previous approaches, and doesn't experience degraded performance until overlap falls below $15\%$. The average performance is summarized in Table~\ref{performanceTable}.

\begin{table}[h]
\centering
\caption{Performance of global registration algorithms on two data sets. Average running time is measured using a single thread on an Intel Core i7-4710MQ CPU clocked at 2.5 GHz.}
\label{performanceTable}
\begin{tabular}{|c|C{1.6cm}|C{1.6cm}|C{1.6cm}|C{1.6cm}|C{1.6cm}|}
\hline
               & PCA & GO-ICP & FPFH & 4PCS & Our Method \\ \hline
Stanford (\%)  & 19.5 & 34.1 & 49.3 & 73.0 & 93.6        \\ \hline
Princeton (\%) & 18.5 & 22.0 & 33.0 & 73.2 & 81.5        \\ \hline
Runtime (sec)  & 0.01  & 25 & 3 & 10 & 0.5           \\ \hline
\end{tabular}
\end{table}

\begin{figure}[h]
\centering
\includegraphics[width=\columnwidth]{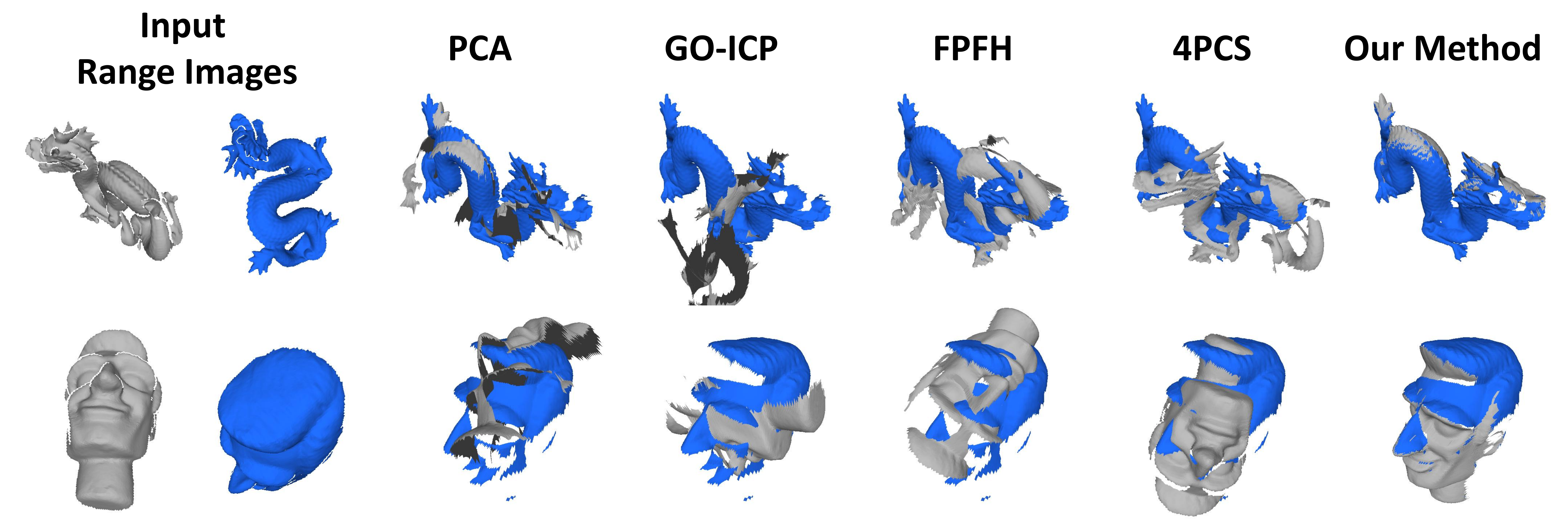}
\caption{Example registration results of range images with limited overlap. First and second row show examples from the Stanford 3D Scanning Repository and the Princeton Shape Benchmark respectively. Please see the supplementary material for more examples.}
\label{fig:FIGRegistrationExamples}
\end{figure}

\begin{figure}[h]
\centering
\includegraphics[width=\columnwidth]{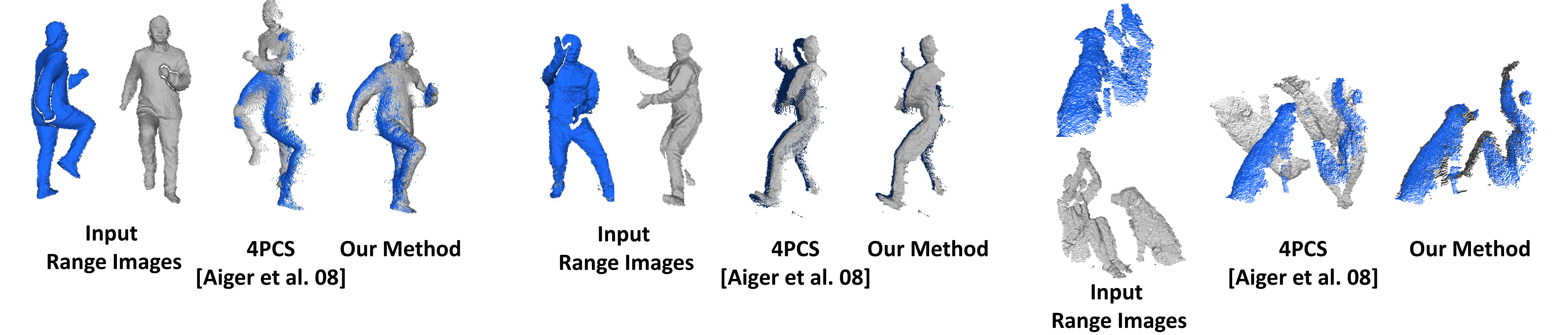}
\caption{Our registration method compared with 4PCS on real data. First two examples are captured by Kinect One sensors while the last example is captured by Structure IO sensors.
\label{fig:realexample}}
\end{figure}

\paragraph{Performance on Real Data.} We further compare the performance of our registration method with 4PCS on pairs of depth maps captured from Kinect One and Structure IO sensors. The hardware setup used to obtain this data is described in detail in the next section. These depth maps share only 10\%-30\% overlap and 4PCS often fails to compute the correct alignment as shown in Figure~\ref{fig:realexample}.

\paragraph{Limitations.} Our global registration method, like most other methods, fails to align scans with dominant symmetries since in such cases depth alone is not enough to resolve the ambiguity. This limitation holds for scans depicting large planar surfaces (e.g. walls and ground) due to continuous symmetry. 

\section{Dynamic Capture Results}
\label{sec:results}


\emph{Hardware.} We provide results of our dynamic scene capture system. We experiment with two popular depth sensors, namely the Kinect One (V2) sensor and the Structure IO sensor. We mount the former on laptops and extend the capture range with long power extension cables. For the latter, we attach it to iPad Air 2 tablets and stream data to laptops through wireless network. Kinect One sensors stream high-fidelity 512x424 depth images and 1920x1080 color images at 30 fps. We use it to cover the entire human body from 3 or 4 views at approximately 2 meters away. Structure IO sensors stream 640x480 for both depth and color (iPad RGB camera after compression) images at 30 fps. Per pixel depth accuracy of the Structure IO sensor is relatively low and unreliable, especially when used outdoor beyond 2 meters. Thus, we use it to capture small objects, \eg, dogs and children, at approximately 1 meter away. Our mobile capture setting allows the subject to move freely in space in stead of being restricted to a specific capture volume.

\emph{Pre-processing.} For each depth image, first we remove background by thresholding depth value and removing dominant planar segments in a RANSAC fashion. For temporal synchronization across depth sensors, we use visual cues, \ie, jumping and clapping hands, to manually initialize the starting frame. Then we automatically synchronize all remaining frames by using the system time stamp of each frame, which is accurate up to milliseconds.


\begin{figure}[t]
\centering
\includegraphics[width=\columnwidth]{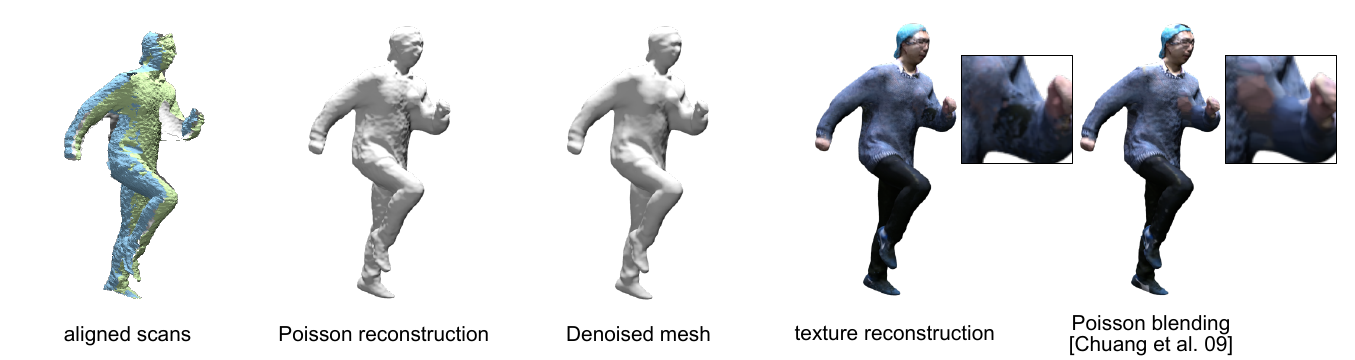}
\caption{From left to right: Globally aligned partial scans from multiple depth sensors; The water-tight mesh model after Poisson reconstruction \cite{kazhdan2006poisson}; Denoised mesh after merging neighboring meshes by using \cite{Li:2012:TCC:2077341.2077343}; Model after our dense correspondences based texture reconstruction; Model after directly applying texture-stitcher \cite{chuang2009estimating}.
\label{fig:textureeval}}
\end{figure}

\emph{Performance.} We process data using a single thread Intel Core i7-4710MQ CPU clocked at 2.5 GHz. It takes on average 15 seconds to globally align all the views for each frame, 5 minutes for surface denoising and reconstruction, and 3 minutes for building dense correspondences and texture reconstruction. 

\begin{figure}[t!]
 \centering
\includegraphics[width=\columnwidth]{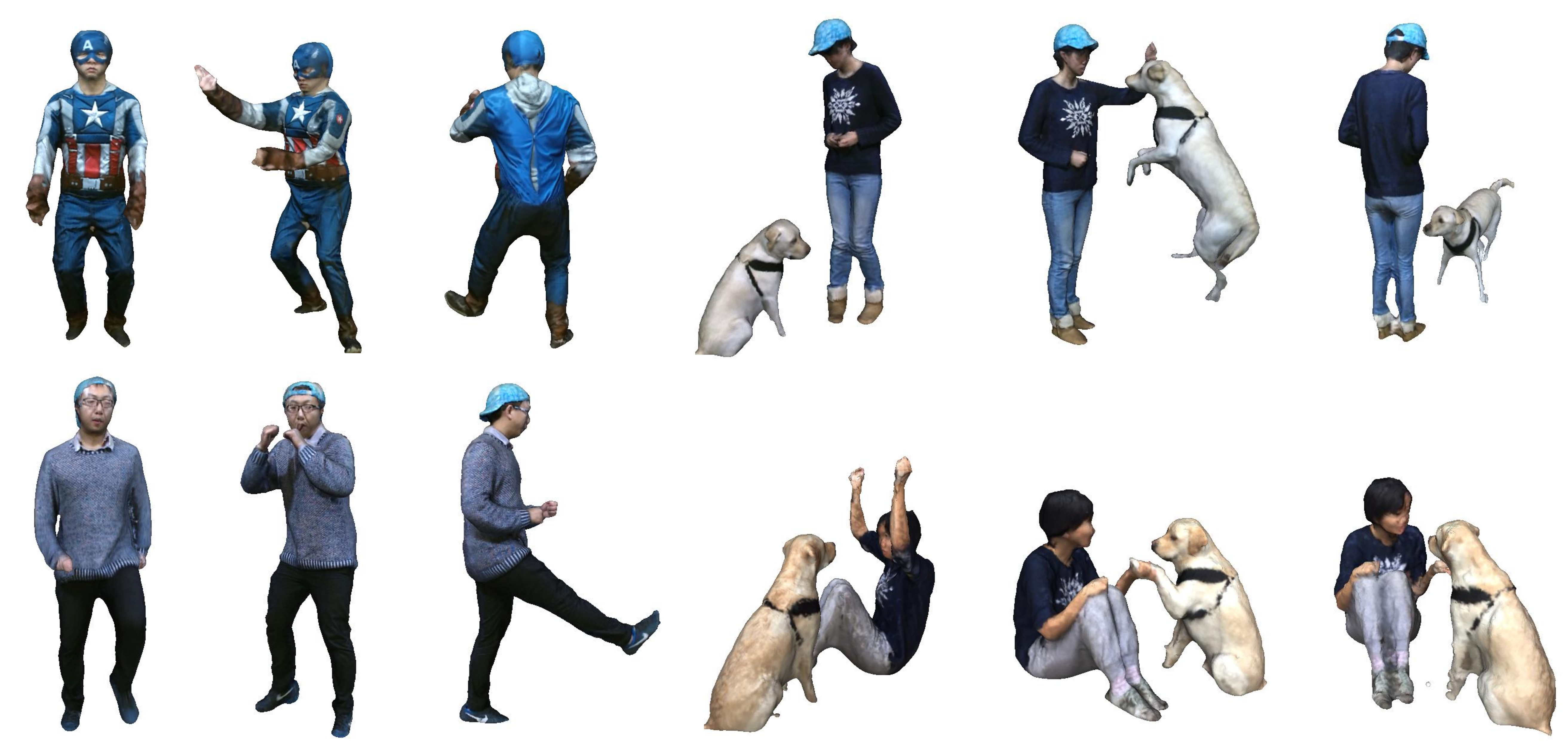}
\caption{Example capturing results. The sequence in the lower right corner is reconstructed from Structure IO sensors, while other sequences are reconstructed from Kinect One Sensors.}
 \label{fig:results2}
\end{figure}

\emph{Results.} We capture a variety of motions and objects, including walking, jumping, playing Tai Chi and dog training (see the supplementary material for a complete list). For all captures, the performer(s) are able to move freely in space while 3 or 4 people follow them with depth sensors. As shown in Figure~\ref{fig:textureeval}, our geometry reconstruction method reduces flickering artifacts of the original Poisson reconstruction, and our texture reconstruction method recovers reliable texture on occluded areas. Figure~\ref{fig:results2} provides several examples that demonstrate the effectiveness and flexibility of our capture system. Our global registration method plays a key role as most range images share only 10\% to 30\% overlap. While we demonstrate successful sequences with 3 depth sensors, using an additional sensor typically improves the reconstruction quality since it provides higher overlap between neighboring views leading to a more robust registration. 

As opposed to most existing free-form surface reconstruction techniques, our method can handle performances of subjects that move through a long trajectory instead of being constrained to a capture volume. Since our method does not require a template, it is not restricted to human performances and can successfully capture animals for which obtaining a static template would be challenging. The global registration method employed for each frame effectively reduces drift for long capture sequences. We can recover plausible textures even in regions that are not fully captured by the sensors using textures from frames where they are visible.

%

\section{Conclusion}
\label{sec:conclusion}

We have demonstrated that it is possible, using only a small number of synchronized consumer-grade handheld sensors, to reconstruct fully-textured moving humans, and without restricting the subject to the constrained environment required by stage setups with calibrated sensor arrays. Our system does not require a template geometry in advance and thus can generalize well to a variety of subjects including animals and small children. Since our system is based on low-cost devices and works in fully unconstrained environments, we believe our system is an important step toward accessible creation of VR and AR content for consumers. Our results depend critically on our new alignment algorithm based on the visibility error metric, which can reliably align partial scans with much less overlap than is required by current state-of-the-art registration algorithms. Without this alignment algorithm, we would need to use many more sensors, and solve the sensor interference problem that would arise. We believe this algorithm is an important contribution on its own, as it represents a significant step forward in global registration.

\bibliographystyle{splncs}
\bibliography{egbib}

\begin{thebibliography}{10}

\bibitem{Collet:2015}
Collet, A., Chuang, M., Sweeney, P., Gillett, D., Evseev, D., Calabrese, D.,
  Hoppe, H., Kirk, A., Sullivan, S.:
\newblock High-quality streamable free-viewpoint video.
\newblock In: {ACM SIGGRAPH}. Volume~34., ACM (July 2015)  69:1--69:13

\bibitem{Ye:2013}
Ye, G., Deng, Y., Hasler, N., Ji, X., Dai, Q., Theobalt, C.:
\newblock Free-viewpoint video of human actors using multiple handheld kinects.
\newblock IEEE Transactions on Cybernetics \textbf{43}(5) (2013)  1370--1382

\bibitem{SCAPE}
Anguelov, D., Srinivasan, P., Koller, D., Thrun, S., Rodgers, J., Davis, J.:
\newblock Scape: Shape completion and animation of people.
\newblock ACM Trans. Graph. \textbf{24}(3) (July 2005)  408--416

\bibitem{aiger20084}
Aiger, D., Mitra, N.J., Cohen-Or, D.:
\newblock 4-points congruent sets for robust pairwise surface registration.
\newblock In: ACM Transactions on Graphics (TOG). Volume~27., ACM (2008) ~85

\bibitem{debevec_light_2012}
Debevec, P.:
\newblock The {Light} {Stages} and {Their} {Applications} to {Photoreal}
  {Digital} {Actors}.
\newblock In: {SIGGRAPH} {Asia}, Singapore (November 2012)

\bibitem{Izadi:2011}
Izadi, S., Kim, D., Hilliges, O., Molyneaux, D., Newcombe, R., Kohli, P.,
  Shotton, J., Hodges, S., Freeman, D., Davison, A., Fitzgibbon, A.:
\newblock Kinectfusion: Real-time 3d reconstruction and interaction using a
  moving depth camera.
\newblock In: {UIST}, New York, NY, USA, ACM (2011)  559--568

\bibitem{Tong:2012}
Tong, J., Zhou, J., Liu, L., Pan, Z., Yan, H.:
\newblock Scanning 3d full human bodies using kinects.
\newblock IEEE TVCG \textbf{18}(4) (April 2012)  643--650

\bibitem{Li:2013}
Li, H., Vouga, E., Gudym, A., Luo, L., Barron, J.T., Gusev, G.:
\newblock 3d self-portraits.
\newblock In: {ACM SIGGRAPH Asia}. Volume~32., ACM (November 2013)
  187:1--187:9

\bibitem{Newcombe_2015_CVPR}
Newcombe, R.A., Fox, D., Seitz, S.M.:
\newblock Dynamicfusion: Reconstruction and tracking of non-rigid scenes in
  real-time.
\newblock In: {IEEE CVPR}. (June 2015)

\bibitem{Dou:2015}
Dou, M., Taylor, J., Fuchs, H., Fitzgibbon, A., Izadi, S.:
\newblock 3d scanning deformable objects with a single rgbd sensor.
\newblock In: {IEEE CVPR}. (June 2015)  493--501

\bibitem{Li:2013:RFA}
Li, H., Yu, J., Ye, Y., Bregler, C.:
\newblock Realtime facial animation with on-the-fly correctives.
\newblock In: {ACM SIGGRAPH}. Volume~32., ACM (July 2013)  42:1--42:10

\bibitem{qian2014realtime}
Qian, C., Sun, X., Wei, Y., Tang, X., Sun, J.:
\newblock Realtime and robust hand tracking from depth.
\newblock In: {IEEE CVPR}, IEEE (2014)  1106--1113

\bibitem{oikonomidis2012tracking}
Oikonomidis, I., Kyriazis, N., Argyros, A.A.:
\newblock Tracking the articulated motion of two strongly interacting hands.
\newblock In: {IEEE CVPR}, IEEE (2012)  1862--1869

\bibitem{Bogo:ICCV:2015}
Bogo, F., Black, M.J., Loper, M., Romero, J.:
\newblock Detailed full-body reconstructions of moving people from monocular
  {RGB-D} sequences.
\newblock (December 2015)  2300--2308

\bibitem{Vlasic:2008}
Vlasic, D., Baran, I., Matusik, W., Popovi\'{c}, J.:
\newblock Articulated mesh animation from multi-view silhouettes.
\newblock In: {ACM SIGGRAPH}. SIGGRAPH '08, New York, NY, USA, ACM (2008)
  97:1--97:9

\bibitem{deAguiar:2008}
de~Aguiar, E., Stoll, C., Theobalt, C., Ahmed, N., Seidel, H.P., Thrun, S.:
\newblock Performance capture from sparse multi-view video.
\newblock In: {ACM SIGGRAPH}, New York, NY, USA, ACM (2008)  98:1--98:10

\bibitem{Li:2009}
Li, H., Adams, B., Guibas, L.J., Pauly, M.:
\newblock Robust single-view geometry and motion reconstruction.
\newblock In: {ACM SIGGRAPH Asia}. SIGGRAPH Asia '09, New York, NY, USA, ACM
  (2009)  175:1--175:10

\bibitem{Zollhofer:2014}
Zollh\"{o}fer, M., Nie{\ss}ner, M., Izadi, S., Rehmann, C., Zach, C., Fisher,
  M., Wu, C., Fitzgibbon, A., Loop, C., Theobalt, C., Stamminger, M.:
\newblock Real-time non-rigid reconstruction using an rgb-d camera.
\newblock In: {ACM SIGGRAPH}. Volume~33., New York, NY, USA, ACM (July 2014)
  156:1--156:12

\bibitem{Wu:2013}
Wu, C., Stoll, C., Valgaerts, L., Theobalt, C.:
\newblock On-set performance capture of multiple actors with a stereo camera.
\newblock ACM Trans. Graph. \textbf{32}(6) (November 2013)  161:1--161:11

\bibitem{Liu2014}
Liu, Y., Ye, G., Wang, Y., Dai, Q., Theobalt, C.:
\newblock Human Performance Capture Using Multiple Handheld Kinects.
\newblock In: Computer Vision and Machine Learning with RGB-D Sensors. Springer
  International Publishing, Cham (2014)  91--108

\bibitem{Wand:2007}
Wand, M., Jenke, P., Huang, Q., Bokeloh, M., Guibas, L., Schilling, A.:
\newblock Reconstruction of deforming geometry from time-varying point clouds.
\newblock In: SGP. SGP '07 (2007)  49--58

\bibitem{Wand:2009:ERN}
Wand, M., Adams, B., Ovsjanikov, M., Berner, A., Bokeloh, M., Jenke, P.,
  Guibas, L., Seidel, H.P., Schilling, A.:
\newblock Efficient reconstruction of nonrigid shape and motion from real-time
  3d scanner data.
\newblock {ACM TOG} \textbf{28}(2) (May 2009)  15:1--15:15

\bibitem{Sussmuth:2008}
S\"{u}{\ss}muth, J., Winter, M., Greiner, G.:
\newblock Reconstructing animated meshes from time-varying point clouds.
\newblock In: SGP. SGP '08 (2008)  1469--1476

\bibitem{Tevs:2012}
Tevs, A., Berner, A., Wand, M., Ihrke, I., Bokeloh, M., Kerber, J., Seidel,
  H.P.:
\newblock Animation cartography\&mdash;intrinsic reconstruction of shape and
  motion.
\newblock {ACM TOG} \textbf{31}(2) (April 2012)  12:1--12:15

\bibitem{Vlasic:2009}
Vlasic, D., Peers, P., Baran, I., Debevec, P., Popovi\'{c}, J., Rusinkiewicz,
  S., Matusik, W.:
\newblock Dynamic shape capture using multi-view photometric stereo.
\newblock In: {ACM SIGGRAPH Asia}. SIGGRAPH Asia '09 (2009)  174:1--174:11

\bibitem{Li:2012:TCC:2077341.2077343}
Li, H., Luo, L., Vlasic, D., Peers, P., Popovi\'{c}, J., Pauly, M.,
  Rusinkiewicz, S.:
\newblock Temporally coherent completion of dynamic shapes.
\newblock {ACM TOG} \textbf{31}(1) (February 2012)  2:1--2:11

\bibitem{TSwET_2012}
Bojsen-Hansen, M., Li, H., Wojtan, C.:
\newblock Tracking surfaces with evolving topology.
\newblock ACM Transactions on Graphics (SIGGRAPH 2012) \textbf{31}(4) (2012)
  53:1--53:10

\bibitem{zhang1994iterative}
Zhang, Z.:
\newblock Iterative point matching for registration of free-form curves and
  surfaces.
\newblock {IJCV} \textbf{13}(2) (1994)  119--152

\bibitem{chen1991object}
Chen, Y., Medioni, G.:
\newblock Object modeling by registration of multiple range images.
\newblock In: ICRA, IEEE (1991)  2724--2729

\bibitem{johnson1999using}
Johnson, A.E., Hebert, M.:
\newblock Using spin images for efficient object recognition in cluttered 3d
  scenes.
\newblock Pattern Analysis and Machine Intelligence, IEEE Transactions on
  \textbf{21}(5) (1999)  433--449

\bibitem{gelfand2005robust}
Gelfand, N., Mitra, N.J., Guibas, L.J., Pottmann, H.:
\newblock Robust global registration.
\newblock In: Symposium on geometry processing. Volume~2. (2005) ~5

\bibitem{rusu2008aligning}
Rusu, R.B., Blodow, N., Marton, Z.C., Beetz, M.:
\newblock Aligning point cloud views using persistent feature histograms.
\newblock In: Intelligent Robots and Systems, 2008 IEEE/RSJ International
  Conference on, IEEE (2008)  3384--3391

\bibitem{rusu2009fast}
Rusu, R.B., Blodow, N., Beetz, M.:
\newblock Fast point feature histograms (fpfh) for 3d registration.
\newblock In: Robotics and Automation, 2009 IEEE International Conference on,
  IEEE (2009)  3212--3217

\bibitem{makadia2006fully}
Makadia, A., Patterson, A., Daniilidis, K.:
\newblock Fully automatic registration of 3d point clouds.
\newblock In: CVPR, 2006 IEEE Conference on. Volume~1., IEEE (2006)  1297--1304

\bibitem{horn1984extended}
Horn, B.K.:
\newblock Extended gaussian images.
\newblock Proceedings of the IEEE \textbf{72}(12) (1984)  1671--1686

\bibitem{chung1998registration}
Chung, D.H., Yun, I.D., Lee, S.U.:
\newblock Registration of multiple-range views using the reverse-calibration
  technique.
\newblock Pattern Recognition \textbf{31}(4) (1998)  457--464

\bibitem{chen1999ransac}
Chen, C.S., Hung, Y.P., Cheng, J.B.:
\newblock Ransac-based darces: A new approach to fast automatic registration of
  partially overlapping range images.
\newblock Pattern Analysis and Machine Intelligence, IEEE Transactions on
  \textbf{21}(11) (1999)  1229--1234

\bibitem{silva2005precision}
Silva, L., Bellon, O.R., Boyer, K.L.:
\newblock Precision range image registration using a robust surface
  interpenetration measure and enhanced genetic algorithms.
\newblock Pattern Analysis and Machine Intelligence, IEEE Transactions on
  \textbf{27}(5) (2005)  762--776

\bibitem{yang2013go}
Yang, J., Li, H., Jia, Y.:
\newblock Go-icp: Solving 3d registration efficiently and globally optimally.
\newblock In: Computer Vision (ICCV), 2013 IEEE International Conference on,
  IEEE (2013)  1457--1464

\bibitem{mellado2014super}
Mellado, N., Aiger, D., Mitra, N.J.:
\newblock Super 4pcs fast global pointcloud registration via smart indexing.
\newblock In: Computer Graphics Forum. Volume~33., Wiley Online Library (2014)
  205--215

\bibitem{580392}
Moezzi, S., Tai, L.C., Gerard, P.:
\newblock Virtual view generation for 3d digital video.
\newblock MultiMedia, IEEE \textbf{4}(1) (Jan 1997)  18--26

\bibitem{Matusik:2000:IVH}
Matusik, W., Buehler, C., Raskar, R., Gortler, S.J., McMillan, L.:
\newblock Image-based visual hulls.
\newblock In: Proceedings of the 27th Annual Conference on Computer Graphics
  and Interactive Techniques. SIGGRAPH '00, New York, NY, USA, ACM
  Press/Addison-Wesley Publishing Co. (2000)  369--374

\bibitem{4155753}
Franco, J., Lapierre, M., Boyer, E.:
\newblock Visual shapes of silhouette sets.
\newblock In: 3D Data Processing, Visualization, and Transmission, Third
  International Symposium on. (June 2006)  397--404

\bibitem{4587696}
Ahmed, N., Theobalt, C., Dobrev, P., Seidel, H.P., Thrun, S.:
\newblock Robust fusion of dynamic shape and normal capture for high-quality
  reconstruction of time-varying geometry.
\newblock In: {IEEE CVPR}. (June 2008)  1--8

\bibitem{Starck:2007:SCP}
Starck, J., Hilton, A.:
\newblock Surface capture for performance-based animation.
\newblock IEEE Comput. Graph. Appl. \textbf{27}(3) (May 2007)  21--31

\bibitem{Wu_ICCV2011}
Wu, C., Varanasi, K., Liu, Y., Seidel, H.P., Theobalt, C.:
\newblock {Shading-based dynamic shape refinement from multi-view video under
  general illumination}, IEEE (November 2011)  1108--1115

\bibitem{hernandez2007silhouette}
Hern{\'a}ndez, C., Schmitt, F., Cipolla, R.:
\newblock Silhouette coherence for camera calibration under circular motion.
\newblock Pattern Analysis and Machine Intelligence, IEEE Transactions on
  \textbf{29}(2) (2007)  343--349

\bibitem{WeiHCVL15}
Wei, L., Huang, Q., Ceylan, D., Vouga, E., Li, H.:
\newblock Dense human body correspondences using convolutional networks.
\newblock In: {IEEE CVPR}, IEEE (2016)

\bibitem{kennedy2010particle}
Kennedy, J.:
\newblock Particle swarm optimization.
\newblock In: Encyclopedia of Machine Learning.
\newblock Springer (2010)  760--766

\bibitem{shoemake1992uniform}
Shoemake, K.:
\newblock Uniform random rotations.
\newblock In: Graphics Gems III, Academic Press Professional, Inc. (1992)
  124--132

\bibitem{shilane2004princeton}
Shilane, P., Min, P., Kazhdan, M., Funkhouser, T.:
\newblock The princeton shape benchmark.
\newblock In: Shape modeling applications, 2004. Proceedings, IEEE (2004)
  167--178

\bibitem{rusinkiewicz2001efficient}
Rusinkiewicz, S., Levoy, M.:
\newblock Efficient variants of the icp algorithm.
\newblock In: 3-D Digital Imaging and Modeling, IEEE (2001)  145--152

\bibitem{rusu20113d}
Rusu, R.B., Cousins, S.:
\newblock 3d is here: Point cloud library (pcl).
\newblock In: Robotics and Automation (ICRA), 2011 IEEE International
  Conference on, IEEE (2011)  1--4

\bibitem{kazhdan2006poisson}
Kazhdan, M., Bolitho, M., Hoppe, H.:
\newblock Poisson surface reconstruction.
\newblock In: Proceedings of the fourth Eurographics symposium on Geometry
  processing. Volume~7. (2006)

\bibitem{chuang2009estimating}
Chuang, M., Luo, L., Brown, B.J., Rusinkiewicz, S., Kazhdan, M.:
\newblock Estimating the laplace-beltrami operator by restricting 3d functions.
\newblock In: Computer Graphics Forum. Volume~28., Wiley Online Library (2009)
  1475--1484

\bibitem{Chuang:2009:ETL}
Chuang, M., Luo, L., Brown, B.J., Rusinkiewicz, S., Kazhdan, M.:
\newblock Estimating the {Laplace}-{Beltrami} operator by restricting {3D}
  functions.
\newblock Symposium on Geometry Processing (July 2009)

\bibitem{Zhou:2014:CMO:2601097.2601134}
Zhou, Q.Y., Koltun, V.:
\newblock Color map optimization for 3d reconstruction with consumer depth
  cameras.
\newblock In: {ACM SIGGRAPH}. Volume~33., ACM (July 2014)  155:1--155:10

\end{thebibliography}


\title{Capturing Dynamic Textured Surfaces \\of Moving Targets \\ \emph{Supplementary Material}} 

\titlerunning{Capturing Dynamic Textured Surfaces of Moving Targets}

\authorrunning{R. Wang et al.}

\author{}

\institute{}

\maketitle

\section*{Supplemental to Section 4.2 -- Particle Update Methods}

This section discusses in detail the update methods for guide particles and regular particles during the particle swam optimization.

\paragraph{Guide Particle Update.} Here we describe how we update guide particle $T_i^k = (R_i^k, \bs{t}_i^k)$ ($i$ is the particle index and $k$ is the current frame number). We parameterize the tangent space of $SE(3)$ at $T^k_i$ by $\bs{m} = [\bs{u},\bs{v}] \in \mathbb{R}^3 \oplus \mathbb{R}^3$ with $exp(\bs{m}) =  \left(\exp([\bs{u}]_{\times})R^k_i, \bs{t}^k_i + \bs{v}\right)$, where $[\mathbf{u}]_\times$ is the cross-product matrix $[\mathbf{u}]_\times \mathbf{w} = \mathbf{u}\times \mathbf{w}$. For any fixed $\bs{m}$, and partial scans $P_1, P_2$, we can separate $\exp(-\bs{m})P_1$ and $\exp(\bs{m})P_2$ into regions $O^j, F^j, B^j, j\in \{1,2\}$, as described in the main text. We then have

\begin{align*}
E(\exp (\bs{m})) &= d\left(\exp (-\bs{m})P_1, P_2\right) + d\left(\exp (\bs{m})P_2, P_1\right) \\
&= \sum\limits_{x\in\mathcal{F}^1}d_{\mathcal{F}}(x,P_2) + \sum\limits_{x\in\mathcal{B}^1}d_{\mathcal{B}}(x,P_2) \\
&+ \sum\limits_{x\in\mathcal{F}^2}d_{\mathcal{F}}(x,P_1) + \sum\limits_{x\in\mathcal{B}^2}d_{\mathcal{B}}(x,P_1) \\
&= \sum\limits_{x\in\mathcal{F}^1}\|x-I_2(x)\|^2 + \sum\limits_{x\in\mathcal{B}^1}\min_{y\in P_2} \left\|\set{P}_{c^2_v}x - \set{P}_{c^2_v} y\right\|^2 \\
&= \sum\limits_{x\in\mathcal{F}^2}\|x-I_1(x)\|^2 + \sum\limits_{x\in\mathcal{B}^2}\min_{y\in P_1} \left\|\set{P}_{c^1_v}x - \set{P}_{c^1_v} y\right\|^2.
\end{align*}

Minimizing $E(\exp(\bs{m}))$ with respect to $\bs{m}$ is then a non-linear least squares problem, which we use Levenberg-Marquardt. We begin with initial guess $\bs{m} = (0,0)$ and iteratively apply the quasi-Newton update

\begin{equation}
\bs{m} \Leftarrow \bs{m}+\Delta\bs{m},
\label{eq:GaussNewton}
\end{equation}

where $\Delta \bs{m}$ is obtained from solving the linear system

\begin{equation}
(\mathbf{J_r}^T\mathbf{J_r}+\lambda\mathbf{I})\Delta\mathbf{m} = -\mathbf{J_r}^T\mathbf{r}.
\label{eq:UpdateRule}
\end{equation}

$\mathbf{r}$ is a stacked column vector such that $E(\exp (\bs{m}))=\mathbf{r}^T\mathbf{r}$, and $\mathbf{J_r}$ is the Jacobian matrix of $\mathbf{r}$ calculated using chain rule. The damping factor $\lambda$ is set as 0.1 throughout all experiments. After m converges, we set $T_i^{k+1}=\exp (\bs{m})$.

\paragraph{Regular Particle Update.} Here we describe how we update regular (non-guide) particle $T_j^k=(R_j^k, \bs{t}_j^k)$ ($j$ is the particle index and $k$ is the current frame number). We parameterize $T_j^k$ as $\bs{p}=[\bs{q}, \bs{t}_j^k]\in \mathbb{R}^3 \oplus \mathbb{R}^3$, where $\bs{q}$ is the imaginary part of the quaternion representation of $R_j^k$. Up to the sign of the real part, which is assumed positive, $\bs{q}$ determines a unique unit quaternion representing the rotation $R_j^k$. $\bs{p}$ is updated in a traditional PSO fashion,

\begin{equation}
\bs{p} \Leftarrow \bs{p}+\omega_p\bs{v}+\omega_b(\bs{b}-\bs{p}) + \omega_g(\bs{g}-\bs{p}),
\end{equation}

where $\bs{v}$ is the velocity of previous iteration, $\bs{b}$ is the best location particle $\bs{p}$ has been at, and $\bs{g}$ is the best particle location within radius $\theta_r$. Please refer to \cite{kennedy2010particle} for more details. The fixed weights $\omega_p$, $\omega_b$ and $\omega_g$ are set as 0.2, 0.3 and 0.3 throughout all experiments. After update, the boundary condition ($\|\bs{q}\|\leq1$) is checked and enforced by normalization if violated.

\section*{Supplemental to Section 3 -- Surface Reconstruction Algorithm}
\label{sec:texture}

This section summarizes the surface reconstruction method. After globally registering partial scans of each frame, we perform Poisson surface reconstruction \cite{kazhdan2006poisson} to fuse three or four partial scans $S_{i,j}$ ($i$ and $j$ denote the frame and the sensor number respectively), and we obtain a sequence of complete, watertight surfaces $\mathcal{W}_1, \mathcal{W}_2, \ldots, \mathcal{W}_M$. To reduce flickering artifacts and to fill holes, we adopt the shape completion pipeline of Li et al~\cite{Li:2012:TCC:2077341.2077343} to warp partial scans from temporally-proximate frames to the current frame geometry. For $\mathcal{W}_i$, we warp $\mathcal{W}_{i-1}$ and $\mathcal{W}_{i+1}$ to align with $W_i$ using a mesh deformation model based on pairwise correspondences and Laplacian coordinates. We further combine them all using Poisson surface reconstruction with the following weights: 10
for the reconstructed mesh of the current frame and the warped neighboring frames, 2 for the hole-filled regions of the current frame, and 1 for the hole-filled regions of the warped neighboring frames. This imposes a mild temporal filter on the reconstructed surfaces, and a strong filter on the hole-filled regions. This step reduces the temporal flicker, and propagates some of the reconstructed surface detail from the neighboring frames onto the current frame (this stems from the neighboring reconstructed mesh
weight being larger than any hole-filled region weight). Please refer to \cite{Li:2012:TCC:2077341.2077343} for more details.

\section*{Supplemental to Section 3 -- Texture Reconstruction Algorithm}
\label{sec:texture}

This section explains in detail the texture reconstruction step based on dense correspondences. After the surface reconstruction step, we first perform texture reconstruction~\cite{Chuang:2009:ETL} to obtain texture for $\mathcal{W}_i$, by fusing and interpolating the texture from partial scans $S_{i,j}$ ($i$ and $j$ denote the frame and the sensor number respectively). However, each surface contains regions where this texture is unreliable, either because the region had poor coverage in the partial scans, or is located near the seam between two partial scans where the texture is inconsistent due to sensor noise and variations in lighting. When capturing clothed humans using three sensors, we observe that roughly 10--20\% of the texture on each surface is unreliable. 

The recent work of Zhou et al~\cite{Zhou:2014:CMO:2601097.2601134} presents impressive results on texturing scanned data. This method, however, assumes that the captured scene is static and thus is not applicable in our setting. Tracking methods like optical flow can be used to transfer texture between consecutive surfaces in our capture sequence, but we found them to be too fragile for our purposes: they fail if the deformation between frames is either too large (so that tracking fails) or too small (so that holes in coverage persist over large numbers of frames). Instead we replace unreliable texture on each surface $\mathcal{W}_i$ by computing dense correspondences between $\mathcal{W}_i$ and other surfaces in the sequence (including temporally distant frames), and transferring texture from surfaces whose texture at the corresponding point is reliable. With this approach we can reconstruct reliable texture even in the presence of large geometry or topology changes over time.

\paragraph{Reliability Weight.} We first need a measure $w_{p}\in [0,1]$ of how reliable the reconstructed texture is at each point $p$ of each surface $\mathcal{W}_i$. Intuitively, texture is most reliable at points that directly face the camera; therefore for partial scans $S_{i,j}$ where p is visible, we set
$$w_{p} = \max\left(0, -n_p \cdot c_v\right)$$
where $n_p$ is the surface normal at $p$ and $c_v$ is the view direction of the sensor that captured $S_{i,j}$. If $p$ is visible in multiple partial scans, we take the maximum weight, and if it is visible in none, we set $w_{p}=0$. Furthermore we feather the weights of points that lie close to the boundaries of any partial scans, as texture at the seams tends to also be unreliable.

\paragraph{Computing Correspondences.} We adopt the method of Wei et al~\cite{WeiHCVL15} to predict a pose-invariant descriptor for every vertex of each $\mathcal{W}_i$. The network of Wei et al is trained on a large dataset of captured and artificial human depth images, and can reliably compute a 16-dimensional unit length descriptor for every vertex, where nearby points in feature space are nearly-corresponding on the surfaces.

\paragraph{Texture Transfer.} We declare all points with $w_{p} < \epsilon$ \emph{unreliable} and all others \emph{reliable}. We set $\epsilon=0.3$ throughout all experiments. We compute descriptors for all reliable points (across all frames) and place them in a KD-tree; for each unreliable point $p$, we compute its 50 nearest neighbors (in feature space) among reliable points, and take as the color of $p$ the weighted average of those neighbors, with each neighbor $q$ weighted by its distance from $p$ in feature space and by $w_q$.

\begin{figure}[t]
\centering
\includegraphics[width=\columnwidth]{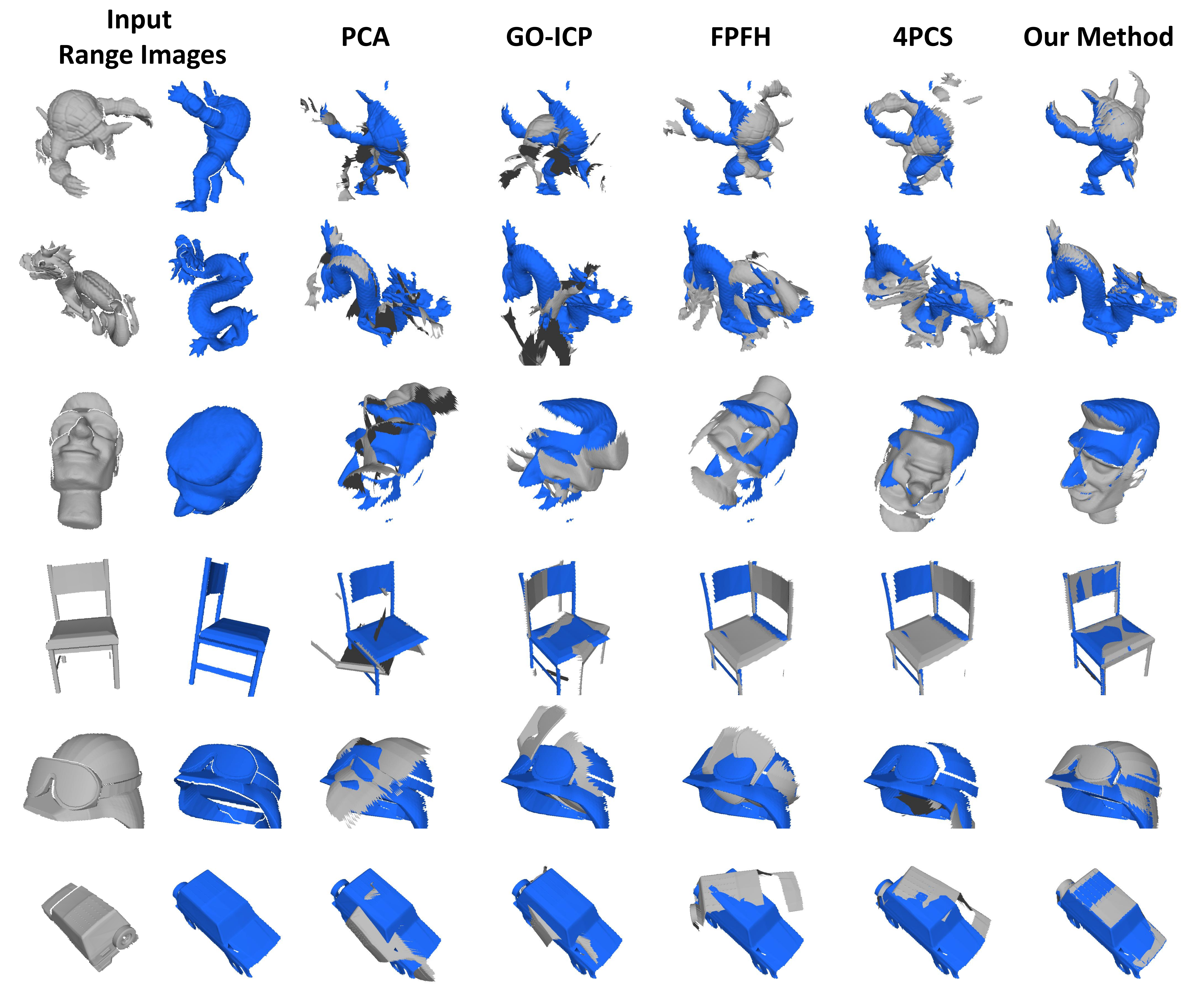}
\caption{Example registration results of range images with limited overlap. First two rows display range image from the Stanford 3D Scanning Repository while the last four rows exhibit data from the Princeton Shape Benchmark.}
\label{fig:FIGRegistrationExamples}
\end{figure}

\section*{Supplemental to Section 5 -- Qualitative Registration Results}

Fig~\ref{fig:FIGRegistrationExamples} below extends figure 7 in the main text, and shows more global registration results.

\section*{Supplemental to Section 6 -- List of Captured Sequence}
This section lists statistics for all captured sequences in Table~\ref{performance}.

\begin{table}[h]
\centering
\caption{\label{performance} List of all captured sequences.}
\begin{tabular}{|c|c|c|c|c|}
\hline
Sequence       & Sensor       & \begin{tabular}[c]{@{}c@{}}Sensor\\ Count\end{tabular} & \begin{tabular}[c]{@{}c@{}}Frame\\ Count\end{tabular} & \begin{tabular}[c]{@{}c@{}}Av. Vertex\\ Count\end{tabular} \\ \hline
Walking 1      & Kinect One   & 3                                                      & 250                                                   & 250,000                                                     \\ \hline
Jumping        & Kinect One   & 3                                                      & 209                                                   & 270,000                                                     \\ \hline
Kicking        & Kinect One   & 3                                                      & 198                                                   & 260,000                                                     \\ \hline
Tai Chi        & Kinect One   & 4                                                      & 491                                                   & 128,000                                                     \\ \hline
Swimming       & Kinect One   & 4                                                      & 370                                                   & 115,000                                                     \\ \hline
Walking 2      & Kinect One   & 4                                                      & 201                                                   & 160,000                                                     \\ \hline
Dog 1 & Kinect One   & 4                                                      & 441                                                   & 150,000                                                     \\ \hline
Dog 2 & Structure IO & 4                                                      & 300                                                   & 145,000                                                     \\ \hline
\end{tabular}
\end{table}

\section*{Supplemental to Section 6 -- Limitations of Capture System}

\begin{figure}[h!]
\centering
\includegraphics[width=\columnwidth]{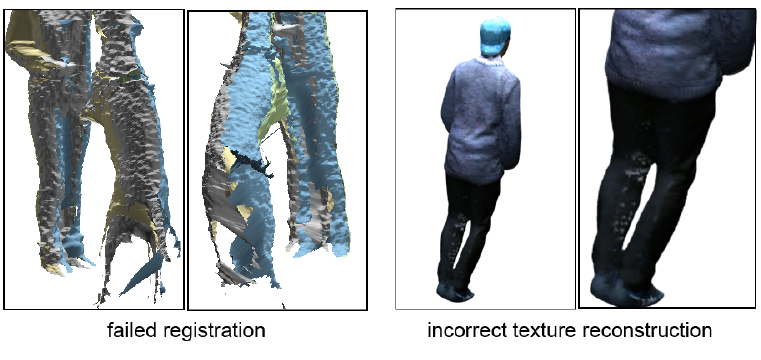}
\caption{Left: Registration failure of frames with fast motion due to minor asynchronization across different sensors. Right: Failed texture reconstruction on consistently occluded regions.}
\label{fig:limitations}
\end{figure}

This section covers limitations of the proposed capture system. The global registration fails when there is barely no overlap, \ie, below $5\%$, potentially caused by two neighboring sensors drifting apart. Our method fails to capture fast motion, \eg, jumping, due to minor asynchronization across different sensors (Figure~\ref{fig:limitations}). Because of the sparse views, there can be potentially consistent occluded regions, for which the texture cannot be accurately recovered from other frames (Figure~\ref{fig:limitations}). Finally, in large occluded regions, Poisson reconstruction might fill in missing surface data with geometry far from the ground truth human shape. In the future we wish to repair these regions by propagating details using a similar approach to how we fix the texture.

\end{document}